\def\year{2018}\relax
\documentclass[letterpaper]{article} %DO NOT CHANGE THIS
\usepackage{aaai18}  %Required
\usepackage{times}  %Required
\usepackage{helvet}  %Required
\usepackage{courier}  %Required
\usepackage{url}  %Required
\usepackage{graphicx}  %Required
\usepackage{times}
\usepackage{xcolor}
\usepackage{soul}
\usepackage[utf8]{inputenc}
\usepackage[small]{caption}
\usepackage{subfigure}

\usepackage{graphicx} % Required for including pictures

\usepackage{float} % Allows putting an [H] in \begin{figure} to specify the exact location of the figure
\usepackage{wrapfig} % Allows in-line images such as the example fish picture

\usepackage{lipsum} % Used for inserting dummy 'Lorem ipsum' text into the template

\usepackage{amssymb}
\usepackage{amsthm}
\usepackage{bm}
\usepackage{amsmath}
\usepackage{enumerate}
\usepackage{url}
\usepackage{algorithm}
\usepackage{algpseudocode}
\title{Iterative Residual Image Deconvolution}

\author{
Li Si-Yao$^1$, 
Dongwei Ren$^2$,
Furong Zhao$^3$
Zijian Hu$^1$,
Junfeng Li$^1$, 
Qian Yin$^1$*
\\ 
$^1$ Beijing Normal University 
$^2$ Tianjin University
$^3$ Sensetime Research 
\\ 
lisiyao@mail.bnu.edu.cn,
\{rendongweihit, hanshzj1998\}@gmail.com,\\
zhaofurong@sensetime.com, 
\{lijunfeng, yinqian\}@bnu.edu.cn}
\begin{document}

\maketitle

\begin{abstract}
 Image deblurring, a.k.a. image deconvolution, recovers a clear image from pixel superposition caused by blur degradation. Few deep convolutional neural networks (CNN) succeed in addressing this task. In this paper, we first demonstrate that the minimum-mean-square-error (MMSE) solution to image deblurring can be interestingly unfolded into a series of residual components. Based on this analysis, we propose a novel iterative residual deconvolution (IRD) algorithm. Further, IRD motivates us to take one step forward to design an explicable and effective CNN architecture for image deconvolution. Specifically, a sequence of residual CNN units are deployed, whose intermediate outputs are then concatenated and integrated, resulting in concatenated residual convolutional network (CRCNet). The experimental results demonstrate that proposed CRCNet not only achieves better quantitative metrics but also recovers more visually plausible texture details compared with state-of-the-art methods.
 \end{abstract}

\section{Introduction}

Image deblurring that aims at recovering a {clear} image from its blurry observation receives considerable research attention in decades.
The blurry image $b$ is usually modeled as a convolution of {clear} image $x$ and blur kernel $k$, i.e.,
\begin{equation} \label{eq:convolution}
b = k * x + \eta,
\end{equation}
where $*$ denotes 2D convolution, and $\eta$ is additive noise.
Thus, image deblurring is also well known as image deconvolution~\cite{andrews1977digital,kundur1996blind}.
When blur kernel $k$ is given, {clear} images can be recovered by deconvolution under the maximum a posterior (MAP) framework~\cite{andrews1977digital,fergus2006removing,levin2007image,krishnan2009fast}:
\begin{equation} \label{eq:map}
\hat{x} = \arg\underset{x}{\min} \left( \|k*x - b\|^2 + \lambda \mathcal{R}(x)\right),
\end{equation}
where $\mathcal{R}(x)$ is regularization term associated with image prior, and $\lambda$ is a positive trade-off parameter.

In conventional deconvolution methods, considerable research attention is paid on the study of regularization term for better describing natural image priors, including Total Variation (TV)~\cite{wang2008new}, hyper-Laplacian~\cite{krishnan2009fast}, dictionary sparsity~\cite{zhang2010bregmanized,hu2010single}, non-local similarity~\cite{dong2013nonlocal}, patch-based low rank prior~\cite{ren2016image} and deep discriminative prior~\cite{li2018learning}.
Note that alternative direction method of multipliers (ADMM) is often employed to efficiently solve these models.
Besides, driven by the success of discriminative learning, the image priors can be learned from abundant training samples.
With the half-quadratic splitting strategy, regression tree field~\cite{jancsary2012regression,schmidt2013discriminative} and shrinkage field~\cite{schmidt2014shrinkage} are proposed to model regularization term, and are effectively trained stage-by-stage.
These learning-based methods have validated the superiority of discriminative learning over manually selected regularization term \cite{schmidt2013discriminative,schmidt2014shrinkage}.

Most recently, deep convolutional neural network (CNN), as a general approximator, has been successfully applied in low level vision tasks, e.g., image denoising~\cite{zhang2017learning}, inpainting~\cite{yang2017high}, supperresolution~\cite{dong2014learning}.
As for image deblurring, there are also several attempts, in which CNN is used to directly map blurry images to clear ones.
In~\cite{nah2017deep}, a deep multi-scale CNN is designed in image deblurring without explicit blur kernel estimation;  
as an upgrade, an recurrent unit is embedded into CNN such that multi scales share same CNN weight~\cite{tao2018}.
In~\cite{Kupyn_2018_CVPR}, a generative adversarial network (GAN) tries to train a ResNet~\cite{he2016deep} supervised by the adversarial discriminator.
However, these trained CNN-based models can only handle mildly blurry images, and usually fail in real cases, since practical blur trajectories are complex.

\begin{figure*}[ht]
\begin{center}
%\fbox{\rule{0pt}{2in} \rule{0.9\linewidth}{0pt}}
   \includegraphics[width=0.95\linewidth, trim={0 160pt 0 120pt}, clip]{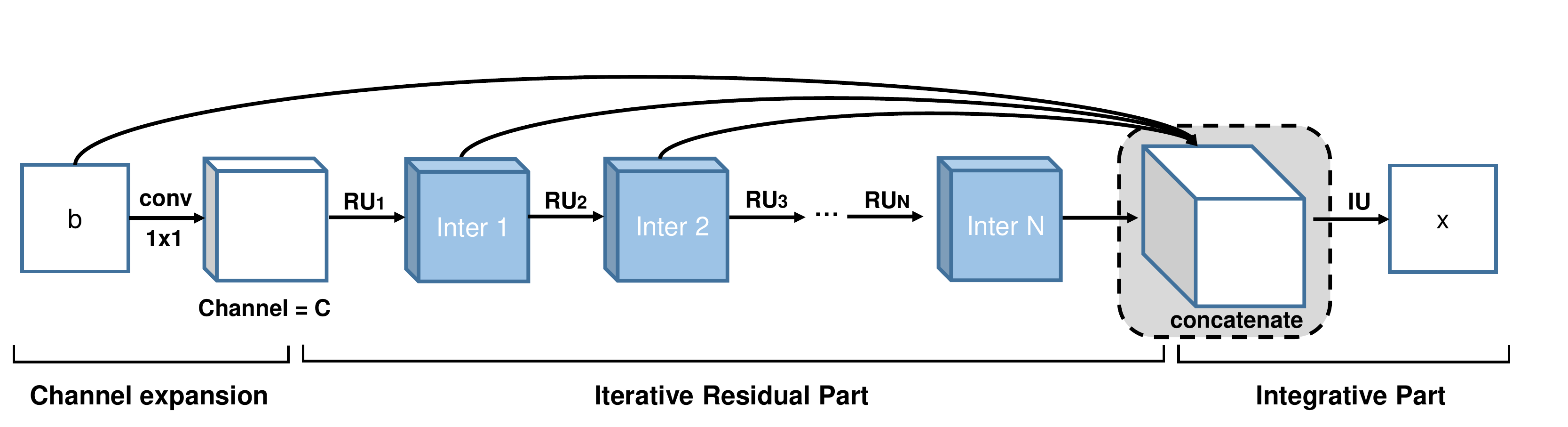}
\end{center}

   \caption{
   The architechture of CRCNet. In this demonstration, $RU_i$ is the $i$-th Residual Unit and $IU$ represents the Integrative Unit.
   }
\label{fig3}
\end{figure*}

When the blur kernel is known, CNN-based deconvolution has also been studied.
On one hand, Xu \textit{et al}~\shortcite{xu2014deep}. have validated that plain CNN cannot succeed in deconvolution.
To make CNN work well for deconvolution, a specific blur kernel would be decomposed into inverse kernels~\cite{xu2014inverse}, which are then used to initialize CNN, inevitably limiting its practical applications.
On the other hand, CNN is incorporated into conventional deconvolution algorithms under plug-and-play strategy.
In~\cite{kruse2017learning}, CNN is employed to solve denoising subproblem in ADMM modulars.
In~\cite{zhang2017learning}, CNN-based Gaussian denoisers are trained off-line, and are iteratively plugged under half-quadratic strategy.
Although these methods empirically achieve satisfactory results, they are not trained end-to-end, and some parameters need to be tuned for balancing CNN strength.
As a summary, the effective CNN architecture for deconvolution still remains unsolved.

\begin{figure}[]
\begin{center}
%\fbox{\rule{0pt}{2in} \rule{0.9\linewidth}{0pt}}
   \includegraphics[width=0.95\linewidth, trim={10pt 250pt 460pt 15pt}, clip]{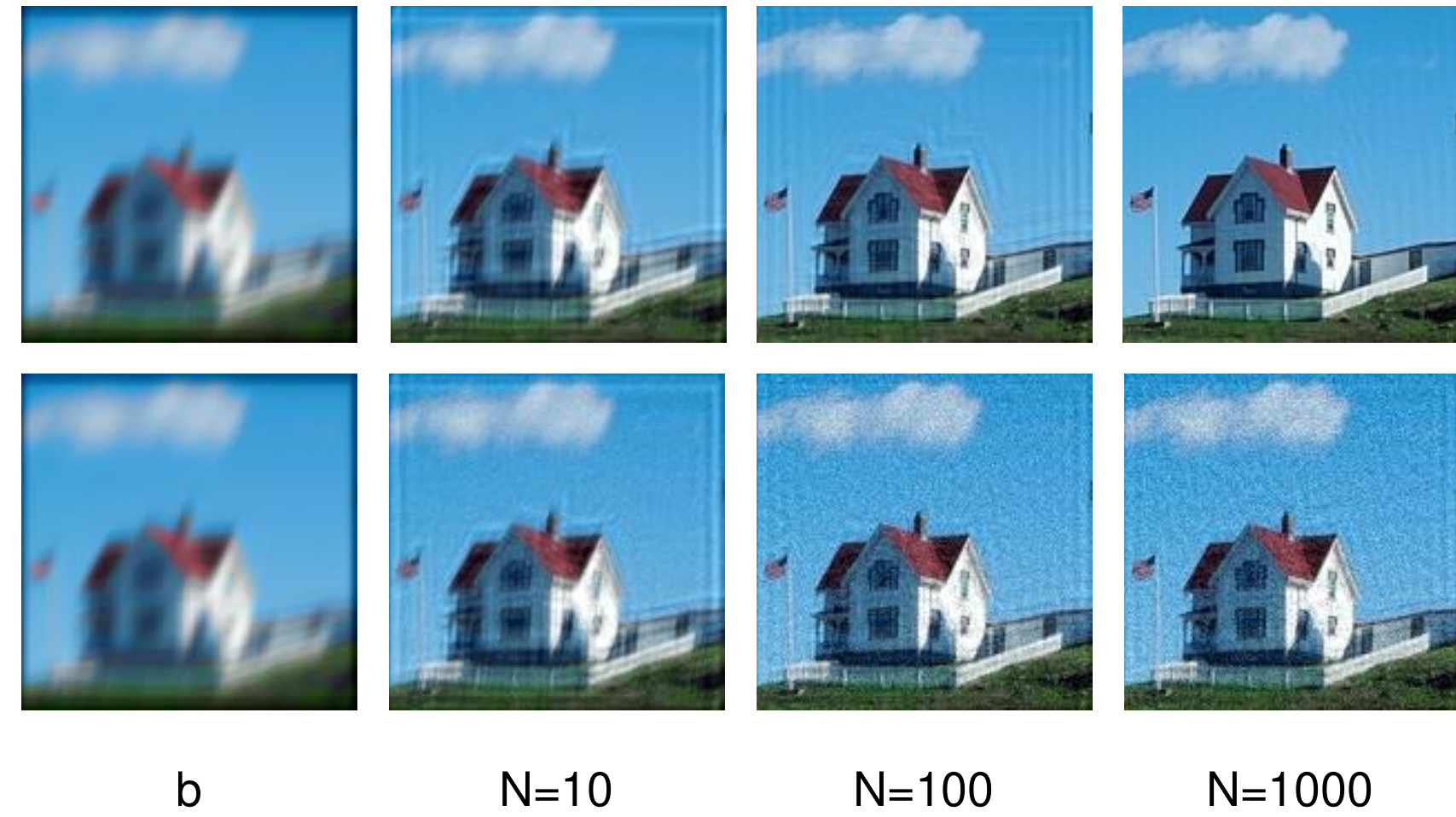}
\end{center}
   \caption{Deblurring results using $ H^T\sum_{n=0}^N(\sigma' I - HH^T)^n$ with different $N$. The first row: deblurring image without noise; the second row: deblurring  noisy image with $\sigma=0.01$. Here pixels of natural images are assumed to be independent and $C_x = I$. For visualization, the deblurring is operated on the first channel of YCbCr.}
\label{fig1}
\end{figure}

In this paper, we propose a novel concatenated residual convolutional network (CRCNet) for deconvolution. 
We first derive a closed-form deconvolution solution driven by the minimum mean square error (MMSE)-based discriminative learning.
Then, using a power series expansion, we unfold MMSE solution into a sum of residual convolutions, which we name iterative residual deconvolution (IRD) algorithm.
IRD is a very simple yet effective deconvolution scheme.
As shown in Figure~\ref{fig1}, the blur can be effectively removed with the increasing of iterations. Although IRD would magnify the noise in degraded image, the blur could still be significantly removed.
Motivated by this observation, we design an effective CNN architecture for deconvolution, as shown in Figure~\ref{fig3}.
We adopt residual CNN unit to substitute the residual component in IRD.
These residual CNN units are iteratively connected, and all intermediate outputs are concatenated and finally integrated, resulting in CRCNet.
Interestingly, the developed non-linear CRC model behaves efficient and robust.
On test datasets, CRCNet can achieve higher quantitative metrics and recover more visually plausible texture details
from  compared with state-of-the-art algorithms. 
We claim that effective CNN architecture plays the critical role in deconvolution, and CRCNet is one of the successful attempts.
Our contributions are three-fold:
\begin{itemize}
\item
  We derive a closed-form deconvolution solution driven by MMSE-based discriminative learning, and further unfold it into a seires, as a simple yet effective IRD algorithm.
\item
  Motivated by IRD algorithm, we propose a novel CRCNet for deconvolution. The CRCNet can be trained end-to-end, but is not a plain CNN architecture and is well analyzed.
\item
  We discuss the contributions of CRCNet, and show the critical role of network architecture for deconvolution. Experimental results demonstrate the superiority of CRCNet over state-of-the-art algorithms.

\end{itemize}

The reminder of this paper is organized as follows: Section 2 derives MMSE-based deconvolution, and then presents IRD algorithm. Section 3 designs CRCNet based on IRD, along with its training strategy. Section 4 demonstrates experimental results and Section 5 ends this paper with conclusions.

\section{Iterative Residual Deconvolution}
%MMSE here
In this section, we first derive a deconvolution solution driven by minimum mean square error (MMSE)~\cite{andrews1977digital}, which is then unfolded via series expansion, resulting in iterative residual deconvolution (IRD) algorithm. Finally we give an insightful analysis to IRD, and provide a potential CNN architecture for deconvolution.

\subsection{MMSE Deconvolution}
The convolution operation in Eqn.~\eqref{eq:convolution} can be equivalently reformatted as a linear transform
\begin{equation}\label{conv2linear}
  \underline{k*x} = H\underline{x},
\end{equation}
where $H$ is a Blocked Toeplitz Matrix~\cite{andrews1977digital} and $\underline{A}$ represents the column-wise expansion vector of matrix $A$.
Then we aim to seek a linear transform $L$ to recover clear image
\begin{equation}\label{linear}
  \hat{\underline{x}} = L\underline{b}.
\end{equation}

Let us assume a set of training image pairs $\{x,b\}$.
By minimizing MSE loss, we have
\begin{equation}\label{linear}
\begin{aligned}
L &=&& \mathrm{arg}\,\min_L \mathbb{E}_{x}[\mathrm{Tr}(\underline{x}-\underline{\hat{x}})(\underline{x}-\underline{\hat{x}})^T]\\
&=&&C_xH^T(HC_xH^T+C_\eta)^{-1},
\end{aligned}
\end{equation}
where $C_x=\underline{x}\underline{x}^T$ and $C_\eta=\underline{\eta}\underline{\eta}^T$ are Gramm matrices of clear images and noises, respectively. $C_x[i, j]$ represents the correlation between the $i$-th pixel and the $j$-th pixel of a sharp image and $C_\eta$ is similar.

% simplified as identity matrix $I$

On one hand, correlations among pixels of natural images are limited~\cite{hu2012psf}, thus eigenvalues of $C_x$ can be deemed to be positive. Then, for any possible $C_x$, we can always find an $\alpha > 0$ such that $\lambda_{max}(\alpha C_x) \le 1$ where $\lambda_{max}$ represents the greatest eigenvalue. Hence, we have
\begin{equation}\label{linear2}
\begin{aligned}
L & = C_xH^T((1/\alpha)(H\alpha C_xH^T + \alpha C_\eta))^{-1}\\
& = C_x'H^T(HC_x'H^T + C_\eta')^{-1},
\end{aligned}
\end{equation}
where $C_x'=\alpha C_x$ and $C_\eta' = \alpha C_\eta$.

On the other hand, $\eta$ is deemed to be zero-meaned gaussian noise with strength $\sigma/\alpha$ (assumed small in this work), thus $C_\eta'$ approximates to $\sigma I$.
Hence, $L$ can be approximated as
\begin{equation}\label{simplified_mmse}
  L \approx C_x'H^T(HC_x'H^T+\sigma I)^{-1}.
\end{equation}
To now, we have obtained a closed-form solution for deconvolution.

\subsection{IRD Algorithm}
For matrix $A$ with $A^n\rightarrow 0$ as $n \rightarrow \infty$, the following series expansion holds:
\begin{equation}\label{series_expansion_A}
  (I-A)^{-1} = \sum_{n=0}^\infty{A^n}.
\end{equation}

As for the case in deconvolution, blur kernel $k$ is under two constraints~\cite{kundur1996blind,levin2009understanding,perrone2014total}:
\begin{equation}
k_{ij} \ge 0
\end{equation}
and
\begin{equation}
  \sum_{i, j} k_{ij} = 1.
\end{equation}
Under such constraints,
the norm of degradation matrix $H$ is limited under 1.
We also found empirically that eigenvalues of $HH^T$ are generally positive.

Thus, the linear deconvolution solution $L$ in Eqn.~\eqref{simplified_mmse} can be unfolded as follows:
\begin{equation}\label{mmse_expansion}
  \boxed{L \approx [C_x'H^T][\sum_{n=0}^N{(\sigma' I - HC_x'H^T)^n]}}
\end{equation}
where $\sigma' = 1 - \sigma$.
\begin{figure}[htb]
\begin{center}
%\fbox{\rule{0pt}{2in} \rule{0.9\linewidth}{0pt}}
   \includegraphics[width=0.95\linewidth, trim={0pt 400pt 475pt 0}, clip]{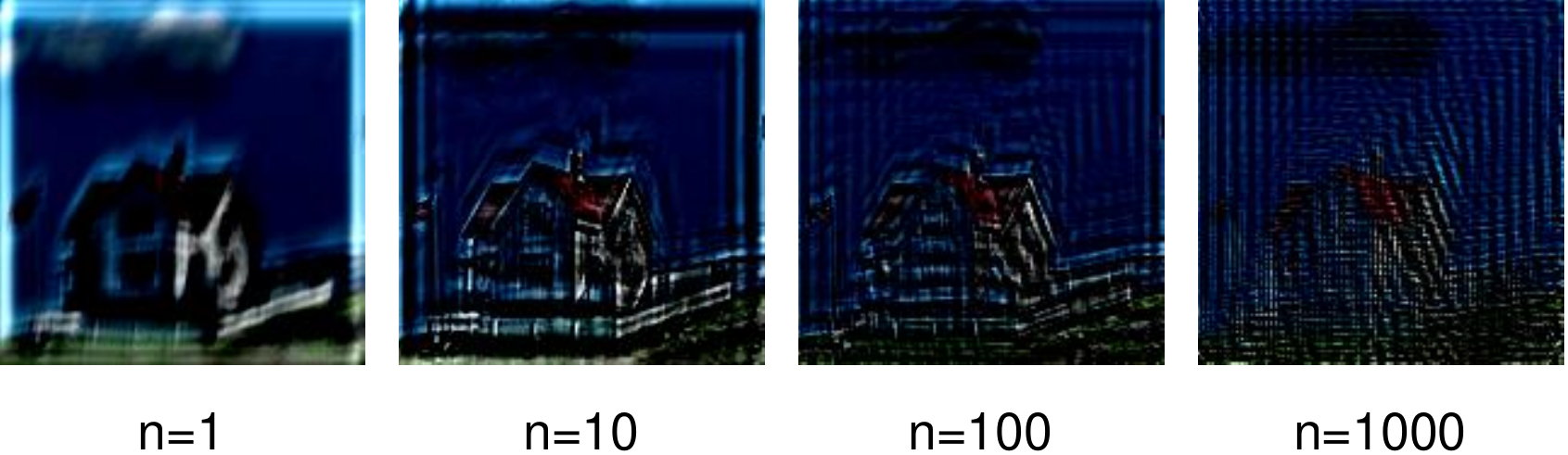}
\end{center}
   \caption{Intermediate components $ H^T(\sigma' I - HH^T)^nb$ corresponding to the first row in Figure~\ref{fig1} with increasing $n$. The operation is taken on the first channel of YCbCr. The original pixel values are too small when $n$ is large, thus images are scaled by $1.5$, $10^2$, $10^3$, $2\times10^4$ respectively to keep the maximum equal to 1 for visualization.}
\label{fig2}
\end{figure}
Eqn.~\eqref{mmse_expansion} can be implemented as an iterative algorithm.
Matrix multiplications by $H$ and $H^T$ are equal to convolutions with $k$ and the flipped kernel $k_-$.\footnote{The flip operation corresponds to $\bm{\text{rot90}}(\cdot, 2)$ in MatLab.}
The correlation matrix $C_x'$ actually plays as the prior of clear images and is assured to be Toeplitz~\cite{andrews1977digital}. Hence, linear transform $C_x'$ is equivalent to a convolution with a limited patch $f_x$ for pixels of clear images are correlated only in the vicinity.
By assuming pixels in a clear image are independent~\cite{hu2012psf,ren2018partial}, $C_x'$ can be simplified as identity matrix $I$, i.e., $f_x=\delta$.
The detailed process is summarized as IRD Algorithm~\ref{algk}.

\begin{algorithm}[t]
\caption{Iterative Residual Deconvolution}\label{algk}
\begin{algorithmic}[1]
\Require{blurry image $b$, degradation kernel $k$, image priori patch $f_x$, noise strength $\sigma$}
\Ensure{restored $\hat{x}$}
\State{$\sigma'\gets1-\sigma$, $s\gets b$, $r\gets b$}
\For{$i\gets0$ to $N - 1$}
\State{$r\gets\sigma'r - k * f_x * k_- * r$}
\State{$s \gets s + r$}
\EndFor{}
\State{$\hat{x} \gets f_x * k_- * s$}

\end{algorithmic}
\end{algorithm}

The IRD algorithm is very simple yet effective for deconvolution.
Figure~\ref{fig1} shows that the clear image can be satisfactorily restored from a noise-free blurry one after 1000 iterations.
Although the noise is significantly magnified for a noisy blurry image, the blur can also be effectively removed.

To explore the significance of unfolded components, we extracted $C_x'H^T(\sigma' I - HC_x'H^T)^n$ as shown in Figure~\ref{fig2}. The energy of iterative residues attenuates but the component represents more detailed signals in higher frequency with increasing $n$. Each iteration extracts residual information from the result of the previous iteration, and those components are finally summed to a clear image.

\subsection{Comparison to Other Unfolded Methods}
The proposed IRD is different from the existing unfolded algorithms. Previous unfolded methods focus on optimization to Eqn.~\eqref{eq:map}. Specifically, ADMM introduces auxiliary variabel $z$ and augmented Lagrange multiplier $y$ into the original object function and optimizes each variable alternately.
As another popular iterative deblur scheme, the accelerated proximal gradient (APG, also named as Fast Iterative Shrinkage/Thresholding Algorithm, FISTA) updates a dual variable to the proximal gradient mapping of \eqref{eq:map}, which significantly accelerates the convergence. A simplified L1-regularized version of ADMM and APG is shown in Algorithm 2 and 3.\footnote{$S_\alpha$ is the soft shrinkage function at $\alpha$}

\begin{algorithm}[t]
\caption{simplified L1-regularized ADMM}\label{algadmm}
\begin{algorithmic}[1]
\Require{blurry image $b$, degradation matrix $H$, trade-off $\rho$}
\Ensure{restored $x$}
\While{not converge}
\State{$ x \gets (H^TH+\rho I)^{-1}(H^Tb + \rho z - y)$}
\State{$ z \gets S_{\lambda / \rho}(x + y/\rho)$}
\State{$ y \gets y + \rho(x - z)$}
\EndWhile{}
\end{algorithmic}
\end{algorithm}

\begin{algorithm}[t]
\caption{simplified L1-regularized APG}\label{algadmm}
\begin{algorithmic}[1]
\Require{blurry image $b$, degradation matrix $H$, trade-off $\rho$}
\Ensure{restored $x$}
\While{not converge}

\State{$ x^{(i)} \gets S_{\lambda t}(y - 2 \lambda t H^T(Hy-b))$}
\State{$ y \gets x^{(i)} + \frac{i - 1}{i + 2}(x^{(i)} - x^{(i-1)})$}
\State{$i \gets i + 1$}
\EndWhile{}
\end{algorithmic}
\end{algorithm}

In contrast, IRD reformulates the inverse process into residual convolutions and represents  MMSE deconvolution as a sum of image components with gradually increasing frequency but lower energy. More interestingly, IRD provides a potential network structure for deconvolution. The iterative residual deconvolution pipeline reminds us the residual learning structure proposed by He \textit{et al.}~\shortcite{he2016deep}. All convolutional parts in IRD can be learned as weights of a CNN. Such an analogy inspired us to propose the following network structure.

\section{Concatenated  Residual Convolutional Network}
\begin{figure}[t]
\begin{center}
%\fbox{\rule{0pt}{2in} \rule{0.9\linewidth}{0pt}}
   \includegraphics[width=\linewidth, trim={160pt 90pt 180pt 80pt}, clip]{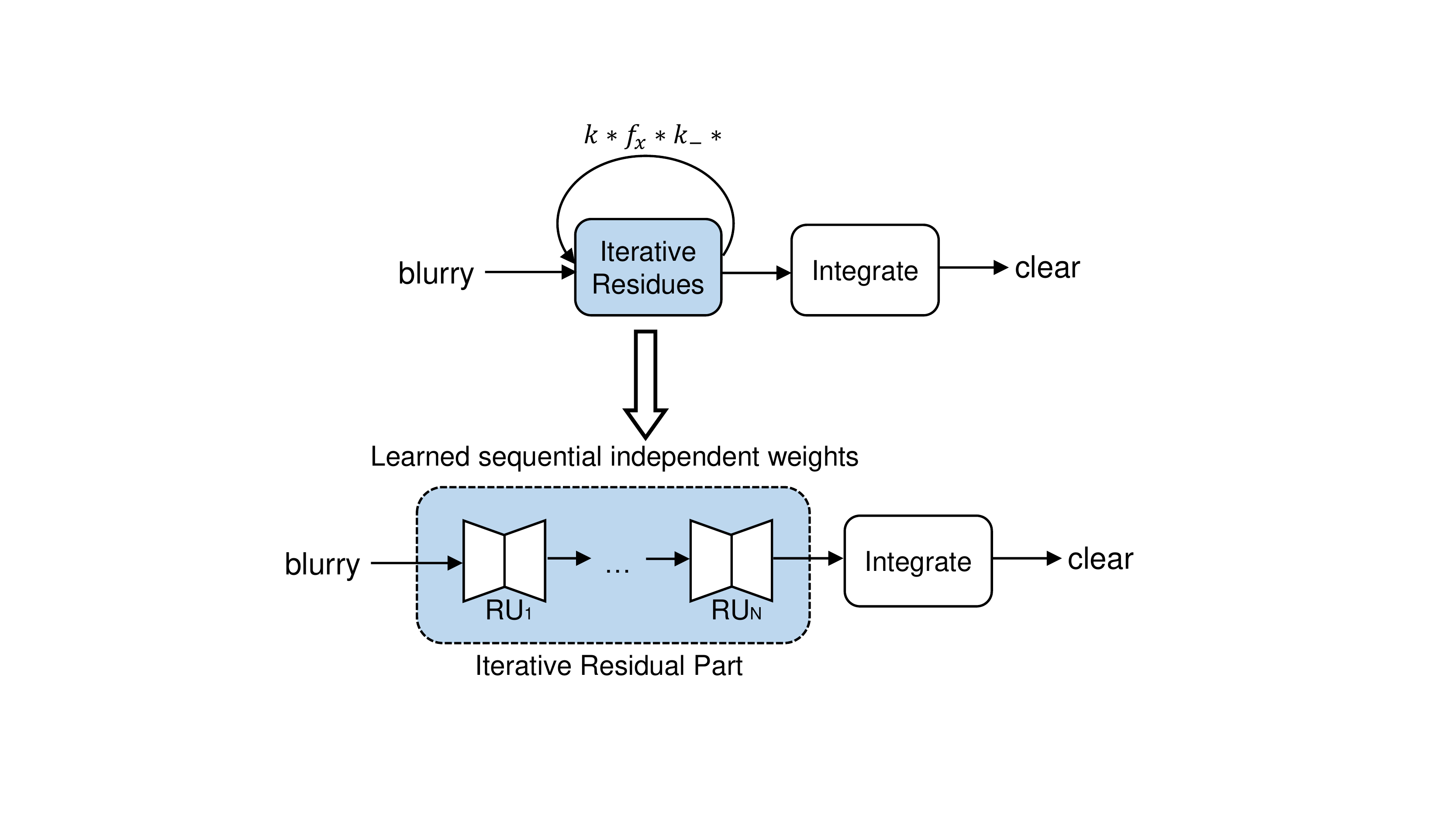}
\end{center}
   \caption{Illustration of IRD to CRCNet: iterative shared weights to squential isolated layers.}
\label{fig:dp}
\end{figure}

By imitating IRD algorithm, we %in this section 
designed a network as shown in Figure~\ref{fig3}, which includes two main parts: the Iterative Residual Part and the Integrative Part, corresponding to $[{(\sigma' I - HC_x'H^T)^n}]$ and $[C_x'H^T\sum_{n=0}^N]$, respectively. 
For the first part, $\sigma'I - HC_x'H^T$ corresponds to the \textit{conv-deconv-minus} structure of a Residual Unit. Considering that linear operator $C_x$ is symmetric $C_x=C_x^{\frac{1}{2}}C_x^{\frac{1}{2}T}$, $HC_xH^T$ can be separated into $(HC_x^{\frac{1}{2}})({C_x^{\frac{1}{2} T}} H^T)$. Note that operator $H$ and $C_x$ are equavalant to convolutions (see section 2), so their transposes correspond to transpose convolutions (also called \textit{deconvolutions} in CNN). 
For the second, operator $[C_x'H^T\sum_{n=0}^N]$ is implemented as \textit{conv} layers on the concatenation of all residues with gradually decreased channels. Because a CNN manipulates convolutions channel-wisely and sum the convolutions of all channels, this structure can sum all residues while adopting convolutions.

We take parametric rectified linear units (PReLU)~\cite{he2015delving} between \textit{conv} or \textit{deconv} layers. The slope of PReLU on negative part is learned during backpropagation,  which can be deemed as a non-linear expansion to IRD.

% \begin{figure}[t]
% \centering
%             \includegraphics[angle=0, width=0.75\linewidth, trim={0pt 250pt 0pt 0pt}, clip]{net_ru_new.pdf}
% \caption{Demonstration of Residual Unit (RU).}
% \label{fig4}
% \end{figure}

% \begin{table}[t]
% \centering
% \begin{tabular}{|c |c |c |} % centered columns (4 columns)
% \hline %inserts double horizontal lines

% Layer & Input/output channels\\
% \hline
% \hline
% \textit{conv}7$\times$7 & 101 / 50\\

% PReLU & 50 / 50\\
% % \hline
% \textit{conv}7$\times$7 & 50 / 25\\
% % \hline
% PReLU & 25 / 25\\
% % \hline
% \textit{conv}7$\times$7 & 25 / 12\\
% % \hline
% PReLU & 12 / 12\\
% % \hline
% \textit{conv}1$\times$1 & 12 / grayscale\\

% \hline %inserts single line
% \end{tabular}

% \caption{Detailed structures of Integration Units (IU) with input/output channels of each layer.} % title of Table
% \label{table:kernel_error} % is used to refer this table in the text
% \end{table}

\paragraph{Channel Expansion.}
As the first step of CRCNet deblurring, the channel of input blurry image is mapped from 1 to $C$ through a 1$\times$1 \textit{conv} layer. The destination of channel expansion is to enhance the capability of \textit{conv/deconv} weights and hence to improve network's flexibility.  

\paragraph{Residual Unit.}

A Residual Unit (RU) calculates the difference between the input and the processed image to extract valid information. 
Formally, 
\begin{equation}\label{ru}
  \begin{aligned}
  RU(x) = x - deconv(PReLU(conv(x))).\\
  \end{aligned}
\end{equation}

Compared to the Eqn.~\eqref{mmse_expansion}, in each RU, convolutional and deconvolutional (transpose convolutional) layers resemble $HC_x'H^T$. However, the auto-encoder-like network can realize more complicated transforms by taking advantage of non-linearity layers. Further, the weights of convolutional layers are  learned from not only the blur but also clear images. Hence, an RU can extract information of images more efficiently.

\paragraph{Iterative Residual Part.}
 The intermediate output of a Residual Unit is fed to the next iteratively. As shown in Figure~\ref{fig3}, $RU_i$ represents the $i$-th RU, and $inter_i$ is the output of $RU_i$. Formally,
\begin{equation}\label{iter_cat}
  \begin{aligned}
  inter_i &=RU_i(inter_{i - 1}),\\
  \end{aligned}
\end{equation}
where $inter_0$ is extended blurry input $b$.

\paragraph{Integration.}

The last part of our network is to integrate all extracted information from the blurry image. The input $b$ and all intermediate residues are concatenated and fed into an Integrative Unit (IU). IU takes three $7\times7$ \textit{conv} layers to play the role of $[C_x'H^T]$ and the channel dimension decreases gradually to 1 through convolutions as a weighted sum of unfolded components.

\begin{figure}[t]
\begin{center}
%\fbox{\rule{0pt}{2in} \rule{0.9\linewidth}{0pt}}
   \includegraphics[width=0.95\linewidth]{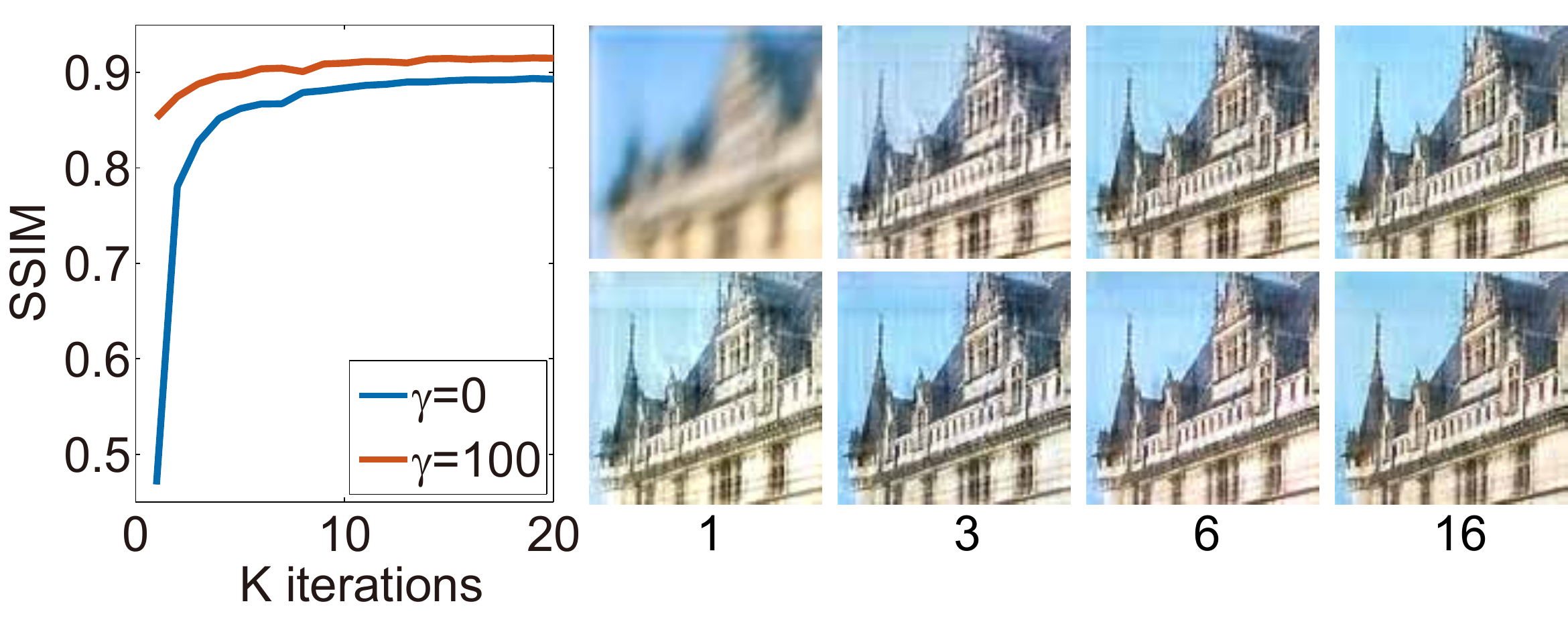}
\end{center}
   \caption{Quantitative and visual comparisons on deblurring a test image with and without taking edge loss. Taking edge loss not only accelerates the training process but also ehances the quality of restored images.}
\label{fig:ssim_epoch_loss}
\end{figure}

\paragraph{Loss Function.}
An ideal deblurring model is expected to restore sufficient content of clear image $x$ and make the restored $\hat{x}$ looks sharp. Thus, the loss function of our network is designed to consist of a \textit{content loss} and an \textit{edge loss}:
\begin{equation}\label{loss_function}
\mathcal{L}=\alpha \mathcal{L}_c+\gamma\mathcal{L}_{e}, 
\end{equation}
where $\mathcal{L}_c$ is the smooth $L_1$ loss~\cite{girshick2015fast}:
\begin{equation}\label{content_loss}
\mathcal{L}_c(\hat{x}, x)= \text{smooth}_{L_1}(\hat{x}- x),
\end{equation}
which is more robust on outliers than MSE loss, and
\begin{equation}\label{edge_loss}
\mathcal{L}_e(\hat{x}, x)= \|\partial_h{\hat{x}}-\partial_hx\|^2+\|\partial_v{\hat{x}}-\partial_vx\|^2,
\end{equation}
in which $\partial_h$ and $\partial_v$ represent horizential and vertical differential operator.

The \textit{edge loss} constrains edges of $\hat{x}$ to be close to those of $x$. Our experiment showed that adding $\mathcal{L}_e$ could speed up the convergence of the network efficiently and make restored edges sharp (See Figure~\ref{fig:ssim_epoch_loss}).

\section{Experimental Results}

\subsection{Training CRCNet}

\subsubsection{Training Dataset Preparation}

{\emph{Clear Image Set}.}
Clear images are essential to train network weights. The dataset is expected to only consist of uniformly sharp and noise-free images with ample textures. We manually selected and clipped 860 $256 \times 256$ RGB images from BSD500~\cite{martin2001database} and COCO~\cite{lin2014microsoft} dataset, during which we omitted all pictures with Bokeh Effect or motion blur.

\begin{figure}[t]
\begin{center}
%\fbox{\rule{0pt}{2in} \rule{0.9\linewidth}{0pt}}
   \includegraphics[width=0.86\linewidth, trim={0pt 70pt 45pt 0}, clip]{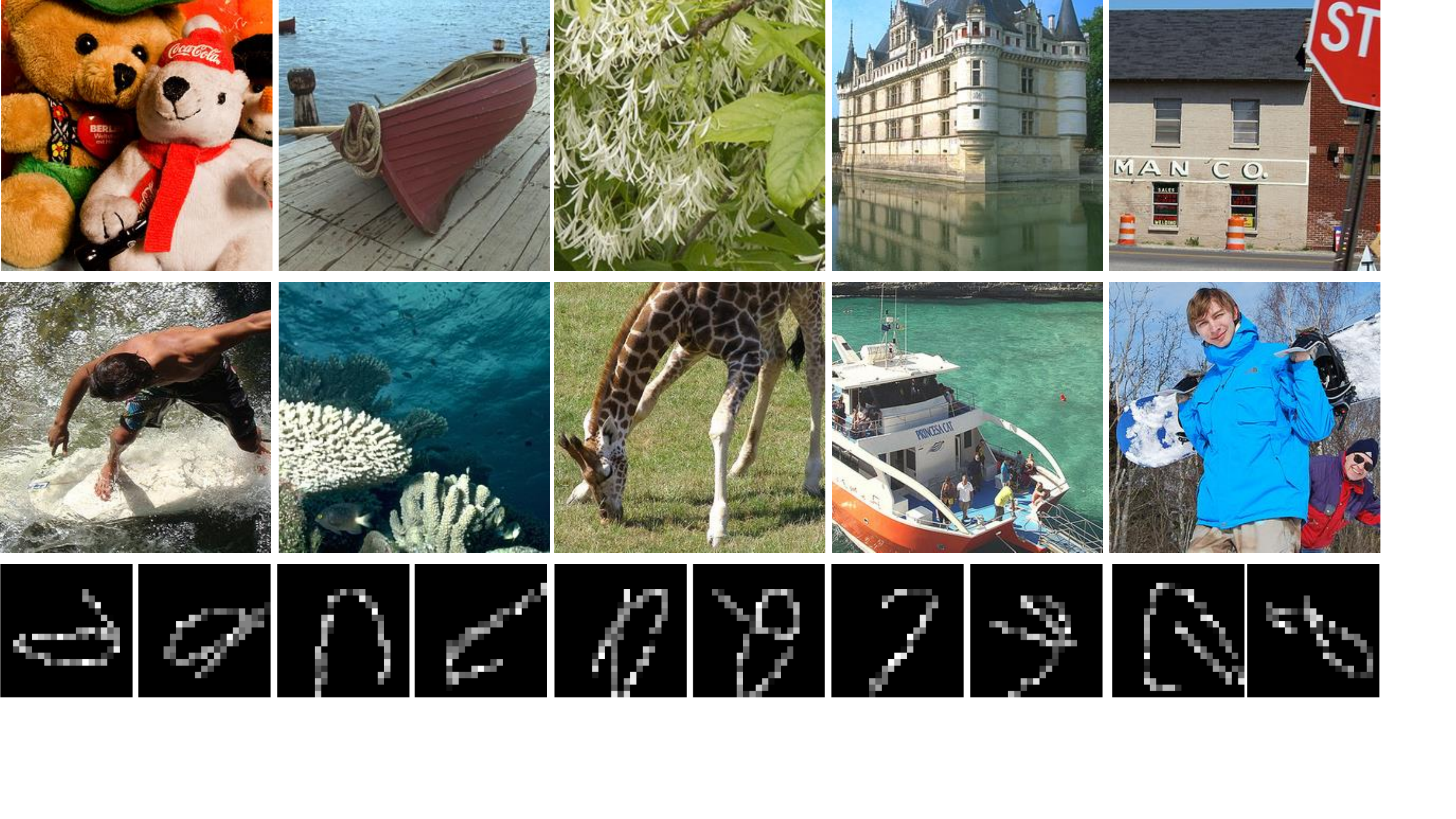}
\end{center}
   \caption{Proposed clear image set and degradation kernels.}
   \label{fig6}
\end{figure}

{\emph{Degradation Kernels}.}
We randomly generated 10 $21\times21$ degradation kernels by using code from~\cite{chakrabarti2016neural} for training and testing. The generated kernels are shown in Figure~\ref{fig6}.

\begin{figure*}[!htb]

\centering
\setlength{\tabcolsep}{1pt}
\small
\begin{tabular}{cclcclcclcclcclccl}

   \multicolumn{3}{c}{\includegraphics[width=.16\linewidth]{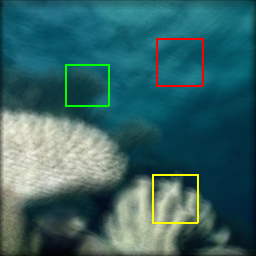}}&
   
   \multicolumn{3}{c}{\includegraphics[width=.16\linewidth]{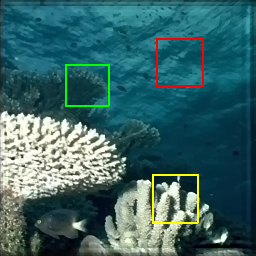}} &
   \multicolumn{3}{c}{\includegraphics[width=.16\linewidth]{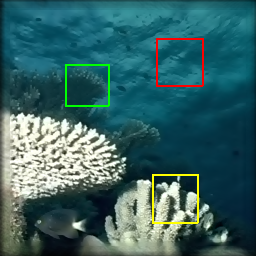}} &
   \multicolumn{3}{c}{\includegraphics[width=.16\linewidth]{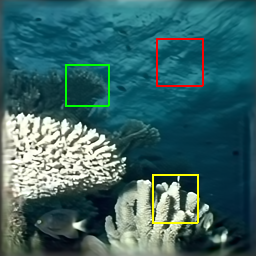}} &
   \multicolumn{3}{c}{\includegraphics[width=.16\linewidth]{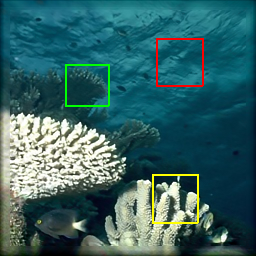}}&
   \multicolumn{3}{c}{\includegraphics[width=.16\linewidth]{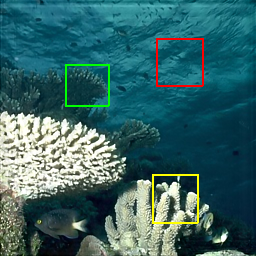}}\vspace{-2pt}\\
   \includegraphics[width=.051\linewidth]{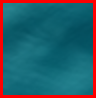}&
   \includegraphics[width=.051\linewidth]{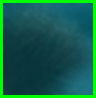}&
   \includegraphics[width=.051\linewidth]{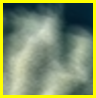}&
   
   \includegraphics[width=.051\linewidth]{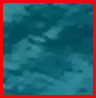}&
   \includegraphics[width=.051\linewidth]{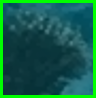}&
   \includegraphics[width=.051\linewidth]{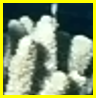}&
   \includegraphics[width=.051\linewidth]{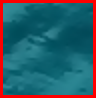}&
   \includegraphics[width=.051\linewidth]{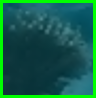}&
   \includegraphics[width=.051\linewidth]{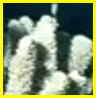}&
   \includegraphics[width=.051\linewidth]{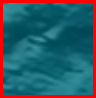}&
   \includegraphics[width=.051\linewidth]{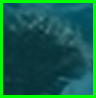}&
   \includegraphics[width=.051\linewidth]{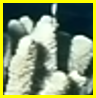}&

   \includegraphics[width=.051\linewidth]{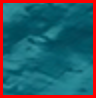}&
   \includegraphics[width=.051\linewidth]{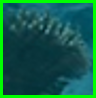}&
   \includegraphics[width=.051\linewidth]{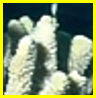}  &

   \includegraphics[width=.051\linewidth]{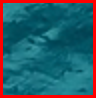}&
   \includegraphics[width=.051\linewidth]{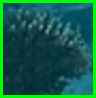}&
   \includegraphics[width=.051\linewidth]{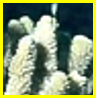}\\
   
   \multicolumn{3}{c}{\includegraphics[width=.16\linewidth]{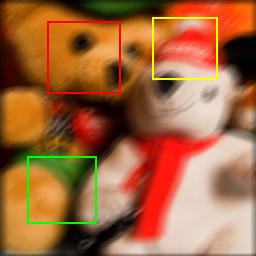}} &
  
   \multicolumn{3}{c}{\includegraphics[width=.16\linewidth]{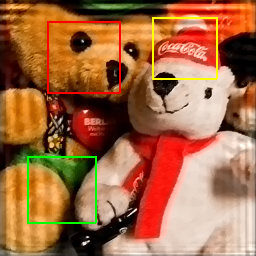}} &
   \multicolumn{3}{c}{\includegraphics[width=.16\linewidth]{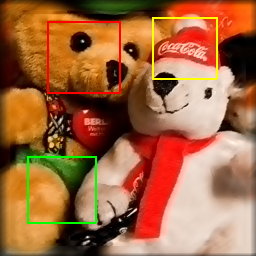}} &
   \multicolumn{3}{c}{\includegraphics[width=.16\linewidth]{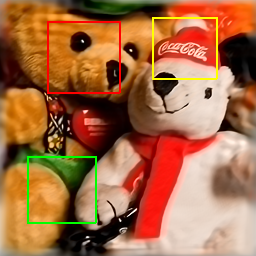}} &

 \multicolumn{3}{c}{\includegraphics[width=.16\linewidth]{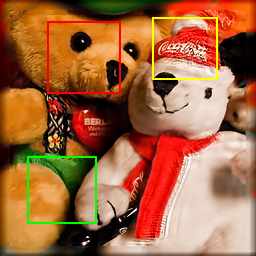}} &

   \multicolumn{3}{c}{\includegraphics[width=.16\linewidth]{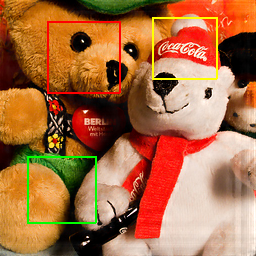}} \vspace{-2pt}\\
   \includegraphics[width=.051\linewidth]{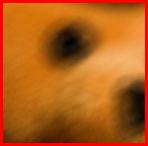}&
   \includegraphics[width=.051\linewidth]{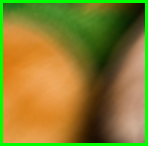}&
   \includegraphics[width=.051\linewidth]{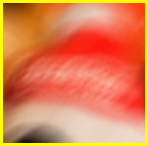} &
  
   \includegraphics[width=.051\linewidth]{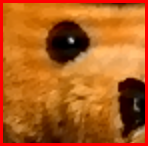}&
   \includegraphics[width=.051\linewidth]{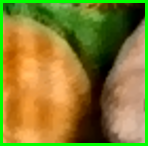}&
   \includegraphics[width=.051\linewidth]{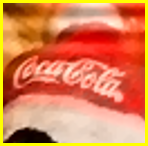}&
   \includegraphics[width=.051\linewidth]{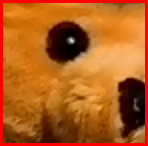}&
   \includegraphics[width=.051\linewidth]{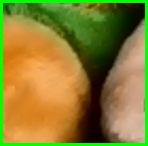}&
   \includegraphics[width=.051\linewidth]{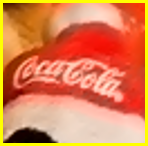}&
   \includegraphics[width=.051\linewidth]{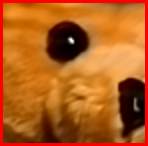}&
   \includegraphics[width=.051\linewidth]{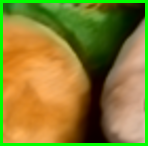}&
   \includegraphics[width=.051\linewidth]{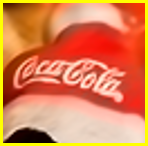}&

   \includegraphics[width=.051\linewidth]{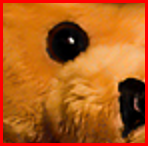}&
   \includegraphics[width=.051\linewidth]{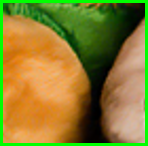}&
   \includegraphics[width=.051\linewidth]{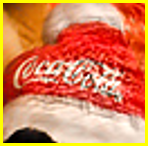}&

   \includegraphics[width=.051\linewidth]{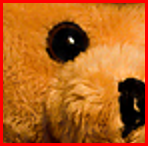}&
   \includegraphics[width=.051\linewidth]{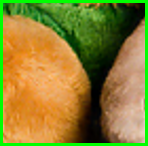}&
   \includegraphics[width=.051\linewidth]{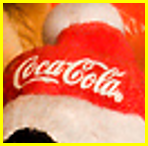}\\

\multicolumn{3}{c}{Blurry images} &
\multicolumn{3}{c}{HL} &
\multicolumn{3}{c}{CSF} &
\multicolumn{3}{c}{IRCNN} &
\multicolumn{3}{c}{FDN} &
\multicolumn{3}{c}{CRCNet}
\end{tabular}
\caption{Visual comparison of deblurring results with state-of-art approaches.}
\label{fig5}
\end{figure*}

\subsubsection{Training details}
We cropped training images into $35\times35$ patches and used Adam~\cite{kingma2014adam} optimizer with a mini-batch of size 32 for training. The initial learning rate is $10^{-4}$ and decay 0.8 per 1000 iterations. The network was only traine 20K iterations for each blur kernel to keep this process portable. In our experiment $\alpha = 5000$ and $\gamma=100$. Ten clear images with ample details were selected for tests and the rest 850 were used for training. 

In our experiments, expanded channel $C=10$. Kernel sizes of $conv$ and $deconv$ of each RU are $7\times7$. We take $N=10$ in iterative residual part and the dimension of the final concatenation before integration is 101.

The CRCNet was implemented in Python with PyTorch and tested on a computer with GeForce GTX 1080 Ti GPU and Intel Core i7-6700K CPU.

\subsection{Comparison to states of the art}
% \begin{figure*}[ht]
% \begin{center}
% %\fbox{\rule{0pt}{2in} \rule{0.9\linewidth}{0pt}}
%    \includegraphics[width=0.92\linewidth, trim={90pt 100pt 70pt 10pt}, clip]{result_may_be_the_last.pdf}
% \end{center}
%    \caption{Visually comparison of deblurring results with state-of-art approaches.}
% \label{fig5}
% \end{figure*}
% table

\subsubsection{Test on synthetic blurry images.}
Quantitative evaluations of average PSNR and SSIM%~\cite{wang2004image} 
of 10 test images with 10 kernels are shown in Table~\ref{table:ssim}. Several test results are shown in Figure~\ref{fig5}. Compared with state-of-art approaches including traditional MAP method using Hyper-Laplacian priors (HL)~\cite{krishnan2009fast}, learning-based method CSF~\cite{schmidt2014shrinkage} and CNN-based methods IRCNN~\cite{zhang2017learning} and FDN~\cite{kruse2017learning}, CRCNet recovers more details in restored images, e.g., fur of Teddy bears and bright spray. Thus deblurred results of our method look more natural and vivid. 

Specifically among contrasts, Schmidt and Roth~\shortcite{schmidt2014shrinkage} substitute the shrinkage mapping of half-quadratic (similar to ADMM but without the Lagrange multiplier) into a learned function constituted by multiple Gaussians, which could be deemed as a learning expansion to gradient-based unfolded method. CRCNet, as a derivative from IRD, achieves better performance. The current literature lacks an learning-expanded deblurring method of APG, thus we don't list relative methods into comparison.

\begin{figure}[t]
\centering
\setlength{\tabcolsep}{1pt}
\small
\begin{tabular}{cccccc}
\multicolumn{1}{c}{\includegraphics[width=.163\linewidth]{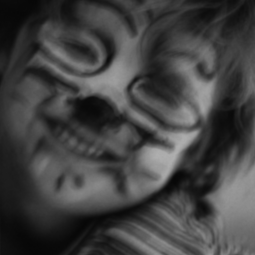}} &

\multicolumn{1}{c}{\includegraphics[width=.163\linewidth]{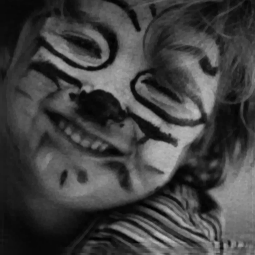}} &
\multicolumn{1}{c}{\includegraphics[width=.163\linewidth]{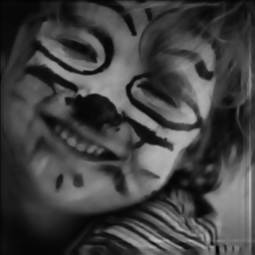}} &
\multicolumn{1}{c}{\includegraphics[width=.163\linewidth]{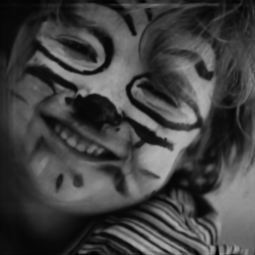}} &
\multicolumn{1}{c}{\includegraphics[width=.163\linewidth]{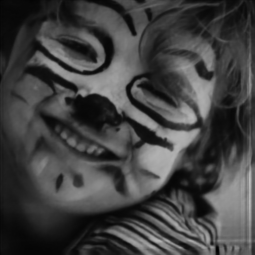}} &
\multicolumn{1}{c}{\includegraphics[width=.165\linewidth]{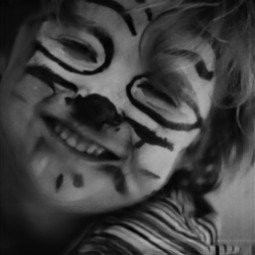}} \vspace{-2pt}\\

\multicolumn{1}{c}{\includegraphics[width=.163\linewidth]{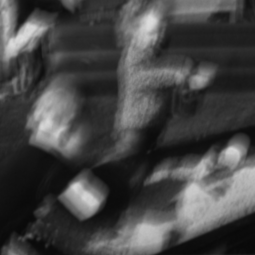}} &
   
   \multicolumn{1}{c}{\includegraphics[width=.163\linewidth]{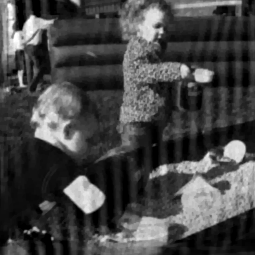}} &
   \multicolumn{1}{c}{\includegraphics[width=.163\linewidth]{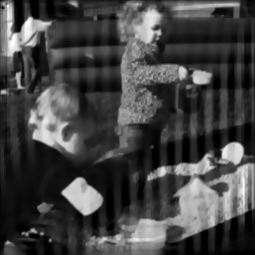}} &
   \multicolumn{1}{c}{\includegraphics[width=.163\linewidth]{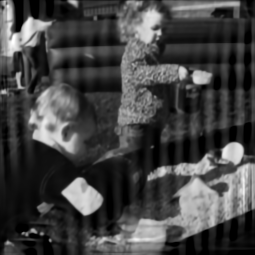}} &
   \multicolumn{1}{c}{\includegraphics[width=.163\linewidth]{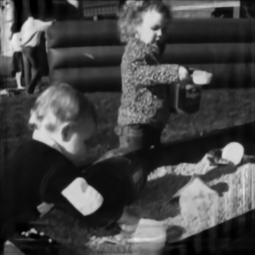}} &
   \multicolumn{1}{c}{\includegraphics[width=.165\linewidth]{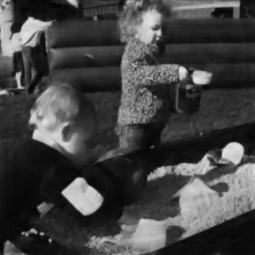}} \vspace{-2pt}\\

   \multicolumn{1}{c}{Blurry} &
\multicolumn{1}{c}{HL} &
\multicolumn{1}{c}{CSF} &
\multicolumn{1}{c}{IRCNN} &
\multicolumn{1}{c}{FDN} &
\multicolumn{1}{c}{CRCNet}\\

\end{tabular}
\caption{Visual comparison of deblurring results on Levin's set with state-of-art approaches.}
\label{fig_deblur_levin}
\end{figure}

We also tested our network on benchmark Levin's set~\cite{levin2009understanding}, which concludes 4 test images $\times$ 8 kernels. The quantitative results are shown in Table~\ref{table:ssim} and visual comparisons are in Figure~\ref{fig_deblur_levin}.
Our approach achieves higher performance both visually and quantitatively.

\subsubsection{Test on real blurry images.}
%  captured GoPro. Blur kernels are estimated by~\cite{zuo2015discriminative}. Direct deconvolution using incorrect kernels results in severe deterioration of restored images (see Figure~\ref{fig_deblur_levins}. Insteed, Deblurring methods have to restore the images based on clear image priors. However, as a trade off, previous methods may sacrifice image details for keeping restored image to be ``clear'' (see Figure~\ref{fig_deblur_levins}). , but achieve relative lower pixel-level distortion scores.
% As we claimed, due to the existence of errors in estimated kernels, deblurring is no longger an inverse problem of convolution. Hence, the comparison on deblurring using inccorect kernels requires visual and perceptual evaluations. Here we take perceptual score proposed in~\cite{ma2017learning} and CRCNet reaches the highest. Much more visual comparisons are shown in supplementary file to convince the reviewers that our method do restore details better visually, even though with a lower distoration score.

We test proposed CRCNet and state-of-the-art methods on real-world blurry images. These blurry images are produced by superposing 16 adjointing frames in motion captured using GoPro Hero 6. Blur kernels are estimated by~\cite{zuo2015discriminative}. Figure~\ref{fig_real} shows that previous methods result in strong ringing effect and hence lower the image quality. In contrast, CRCNet remains plausible visual details while avoiding artifacts. We also take a quantitative perceptual scores proposed in~\cite{ma2017learning} on all methods; CRCNet obtains the highest (see Table \ref{table:ssim_real}). 

%As a note, we add random noise with $\sigma=0.002$ into blur images and kernels during training to improve the generalization of trained CRCNet. 

\begin{table}[t]
\centering % used for centering table
\small
\begin{tabular}{c c c } % centered columns (4 columns)
\hline %inserts double horizontal lines

 Test set  &  Levin  &  Proposed \\

\hline
HL  & 30.20 / 0.90 & 23.56 / 0.77\\
CSF        & 33.53 / 0.93 & 27.24 / 0.85\\
IRCNN      & 34.68 / 0.94 & 27.28 / 0.84\\
FDN        & 35.08 / \textbf{0.96} &29.37 / 0.89 \\
CRCNet     & \textbf{35.39} / \textbf{0.96} & \textbf{29.83} / \textbf{0.92}\\

\hline %inserts single line
\end{tabular}
\caption{Quantitative evaluations of different deblurring methods using ground truth kernel on corresponding test set. Each number pair notes the mean PSNR/SSIM scores.} % title of Table
\label{table:ssim} % is used to refer this table in the text

\end{table}

\begin{table}[t]
\centering % used for centering table
\small
\begin{tabular}{c c c } % centered columns (4 columns)
\hline %inserts double horizontal lines

Method & DCNN & CRCNet \\
\hline

PSNR / SSIM      & 26.77 / 0.85 & 27.05 / 0.86 \\
\hline
param. amount &  15M &   0.4M \\
\hline %inserts single line
\end{tabular}
\caption{Quantitative evaluations of DCNN and CRCNet.} % title of Table
\label{table:ssim_dcnn} % is used to refer this table in the text

\end{table}

\begin{figure}[t]
\centering
\setlength{\tabcolsep}{1pt}
\small
\begin{tabular}{ccc}
\multicolumn{1}{c}{\includegraphics[width=.32\linewidth]{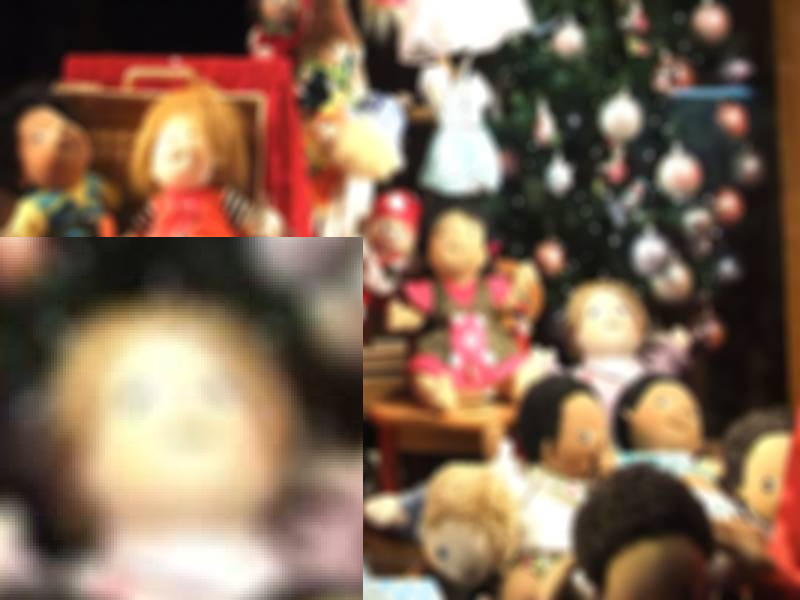}} &

\multicolumn{1}{c}{\includegraphics[width=.32\linewidth]{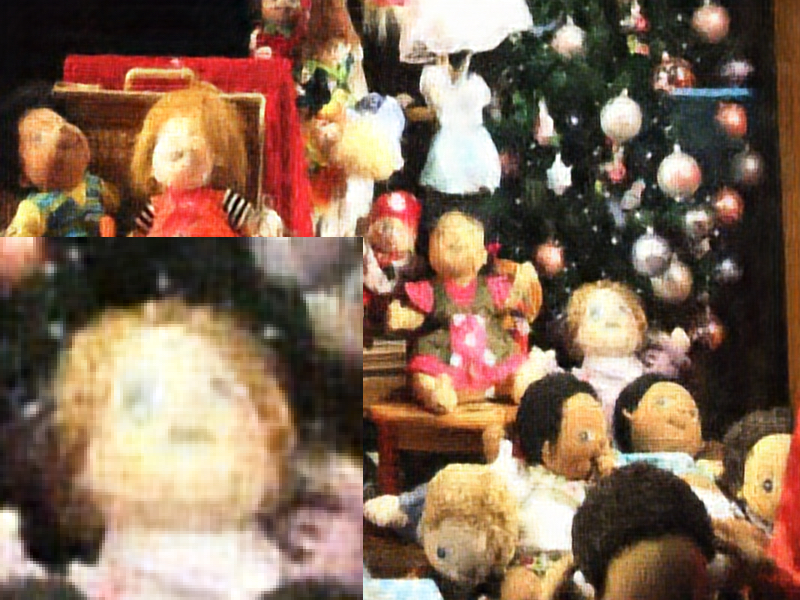}} &
\multicolumn{1}{c}{\includegraphics[width=.32\linewidth]{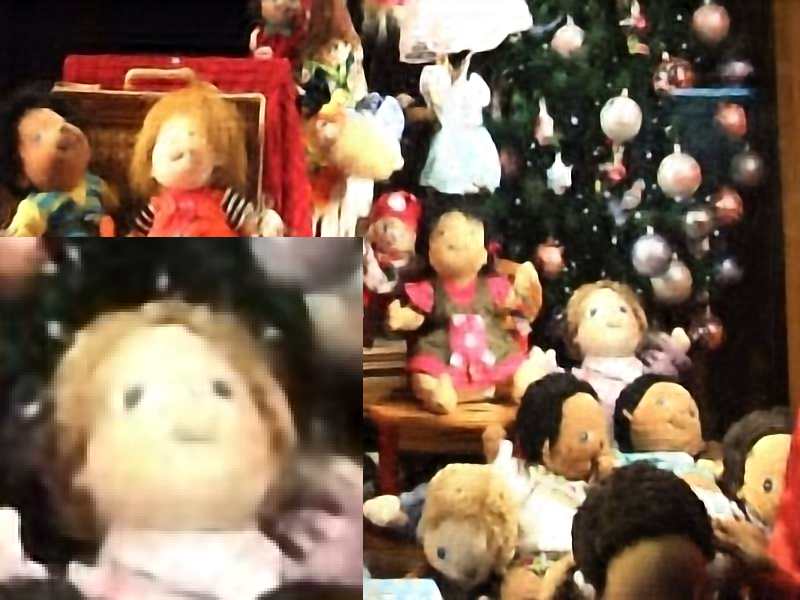}}  \vspace{-2pt}\\

   \multicolumn{1}{c}{Blurry} &
\multicolumn{1}{c}{DCNN} &
\multicolumn{1}{c}{CRCNet}\\
\end{tabular}
\caption{Deblurring \texttt{disk7} examples using DCNN and CRCNet.}
\label{fig_deblur_disk7}
\end{figure}

\subsubsection{Comparison with DCNN.}
In the last part of experiments, we compare our method with previous non-blind deconvolution network DCNN on accompanying dataset in~\cite{xu2014deep} (see Table~\ref{table:ssim_dcnn} and Figure~\ref{fig_deblur_disk7}). This comparison is listed seperatedly for only test code and trained weights on \texttt{disk7} of DCNN are published. In this experiment, kernel is limitted as uniform disk of 7-pixel radius and blurry images are extra degraded by saturation and lossy compression. We also list weight amounts of both networks. CRCNet obtains higher performance while taking much less network parameters. Further, DCNN requires specific initializations while CRCNet can be trained directly in end-to-end way.

The implementation of this work and the clear image set are published at \url{https://github.com/lisiyaoATbnu/crcnet}.

\begin{figure*}[t]

\centering
\setlength{\tabcolsep}{1pt}
\small
\begin{tabular}{cclcclcclcclcclccl}
   \multicolumn{3}{c}{\includegraphics[width=.16\linewidth]{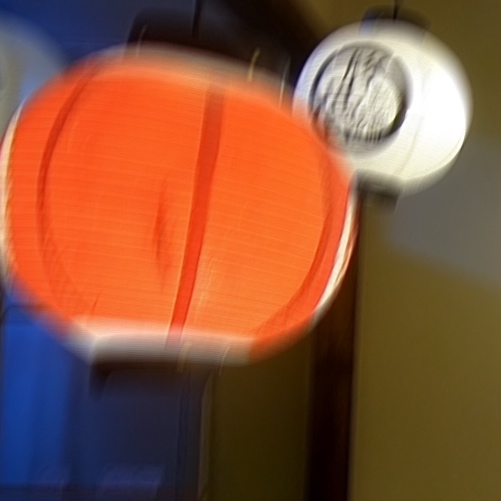}}&
   
   \multicolumn{3}{c}{\includegraphics[width=.16\linewidth]{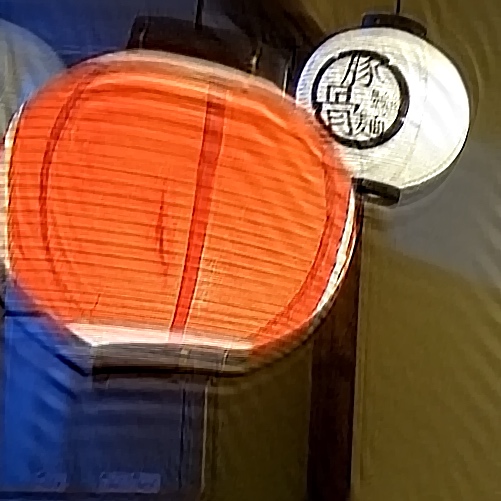}}&
   \multicolumn{3}{c}{\includegraphics[width=.16\linewidth]{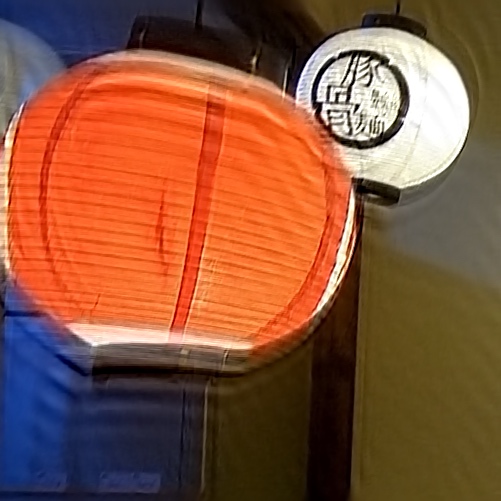}} &
   \multicolumn{3}{c}{\includegraphics[width=.16\linewidth]{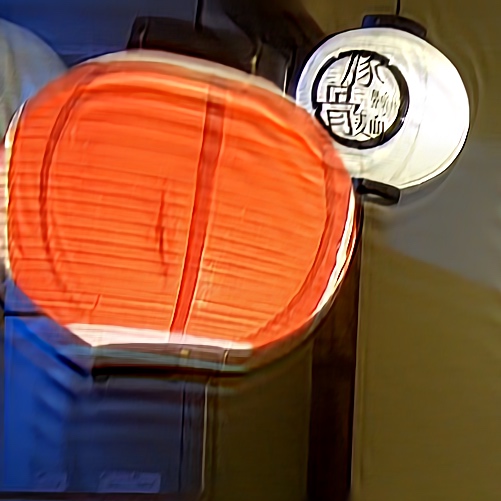}} &
   \multicolumn{3}{c}{\includegraphics[width=.16\linewidth]{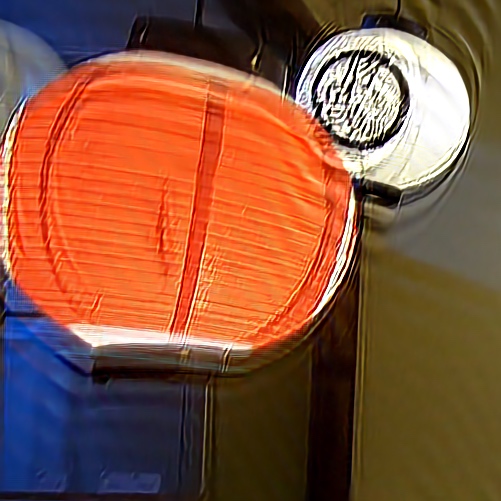}} &
   \multicolumn{3}{c}{\includegraphics[width=.16\linewidth]{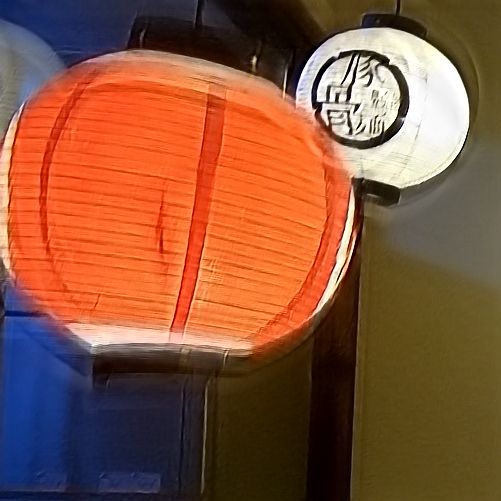}} \vspace{-2pt}\\

    \multicolumn{3}{c}{\includegraphics[width=.16\linewidth]{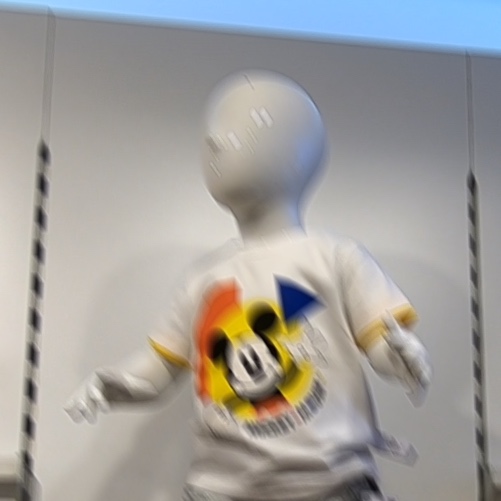}}&
   
   \multicolumn{3}{c}{\includegraphics[width=.16\linewidth]{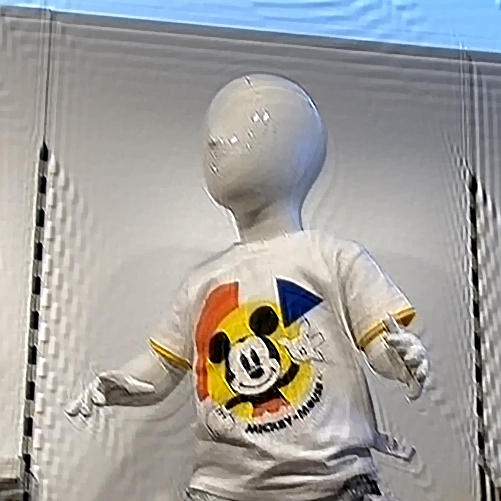}}&
   \multicolumn{3}{c}{\includegraphics[width=.16\linewidth]{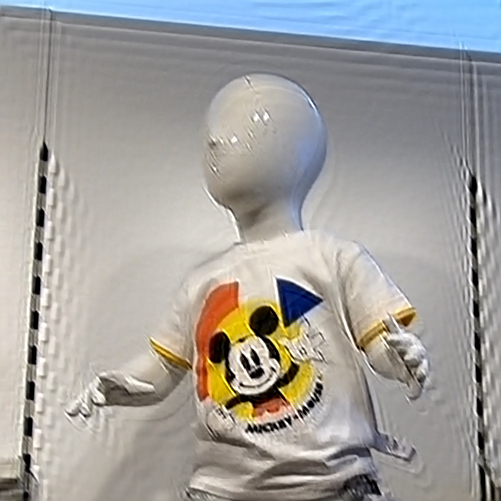}} &
   \multicolumn{3}{c}{\includegraphics[width=.16\linewidth]{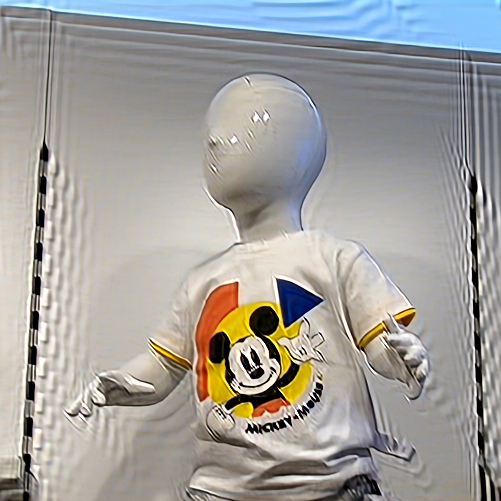}} &
   \multicolumn{3}{c}{\includegraphics[width=.16\linewidth]{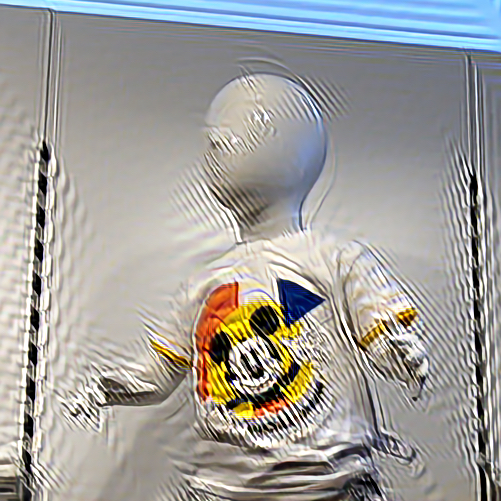}} &
   \multicolumn{3}{c}{\includegraphics[width=.16\linewidth]{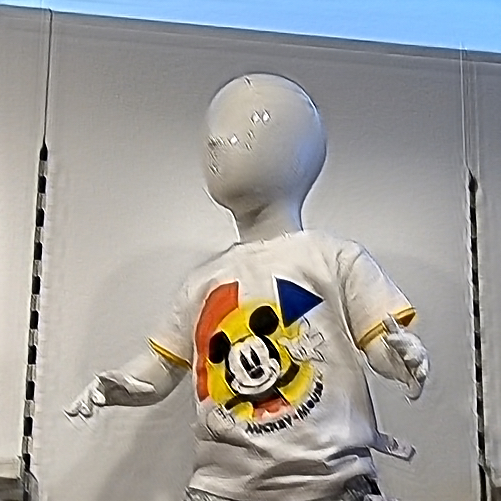}} \vspace{-2pt}\\
   
\multicolumn{3}{c}{\includegraphics[width=.16\linewidth]{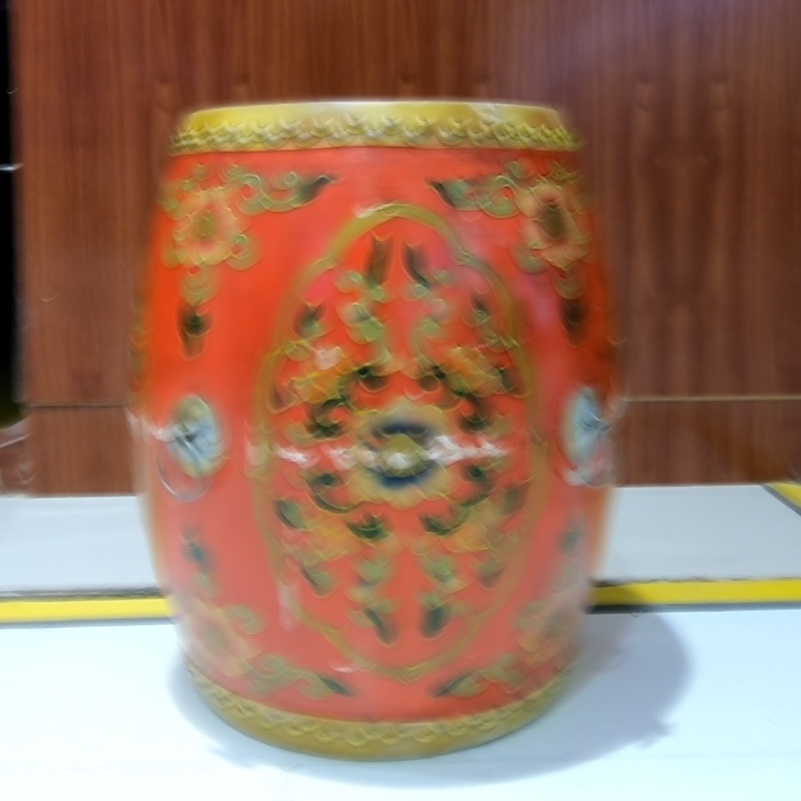}}&
   
   \multicolumn{3}{c}{\includegraphics[width=.16\linewidth]{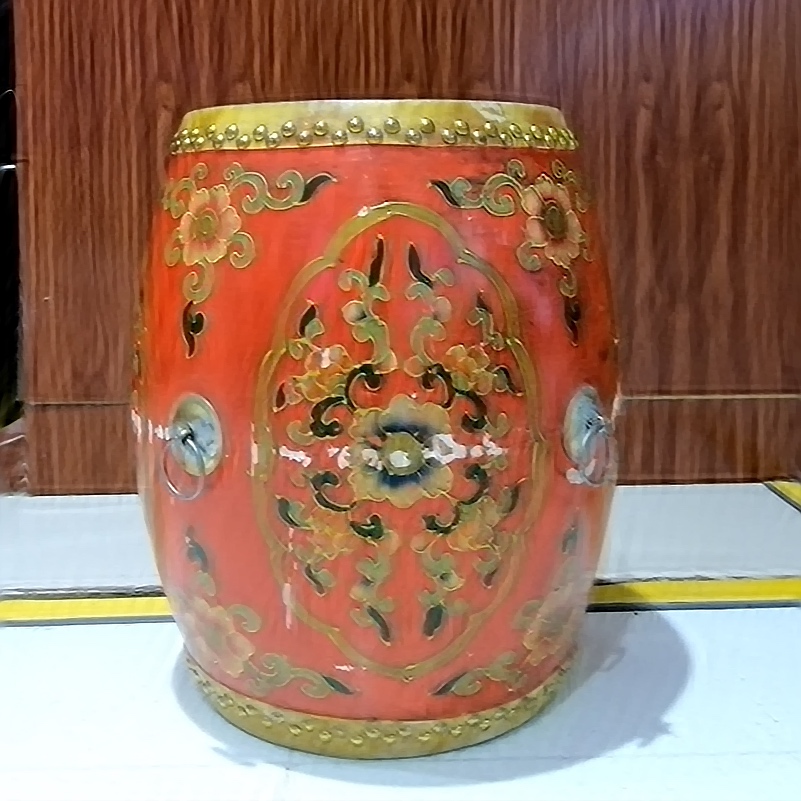}}&
   \multicolumn{3}{c}{\includegraphics[width=.16\linewidth]{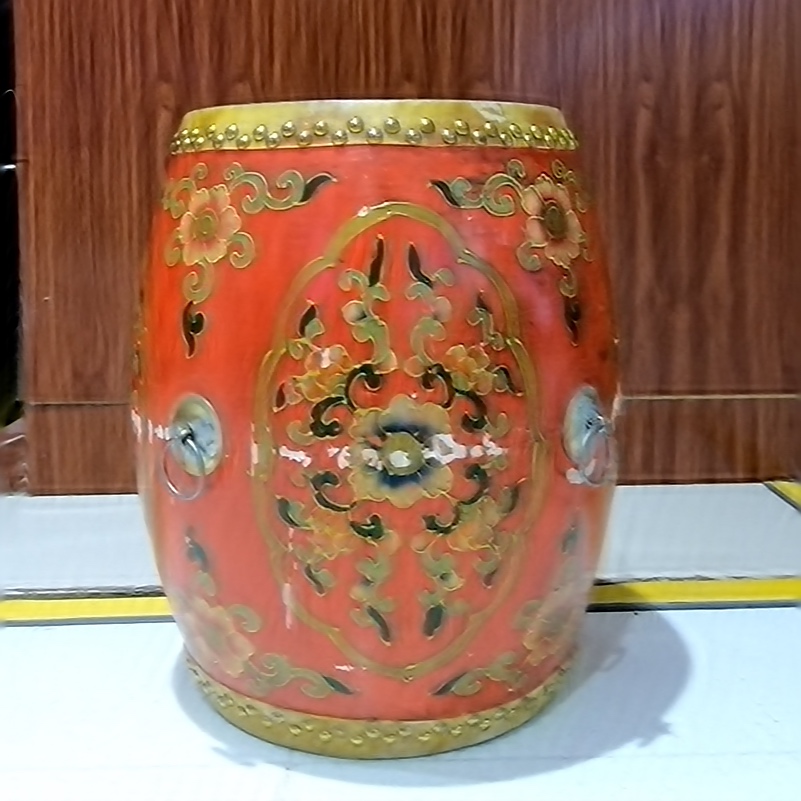}} &
   \multicolumn{3}{c}{\includegraphics[width=.16\linewidth]{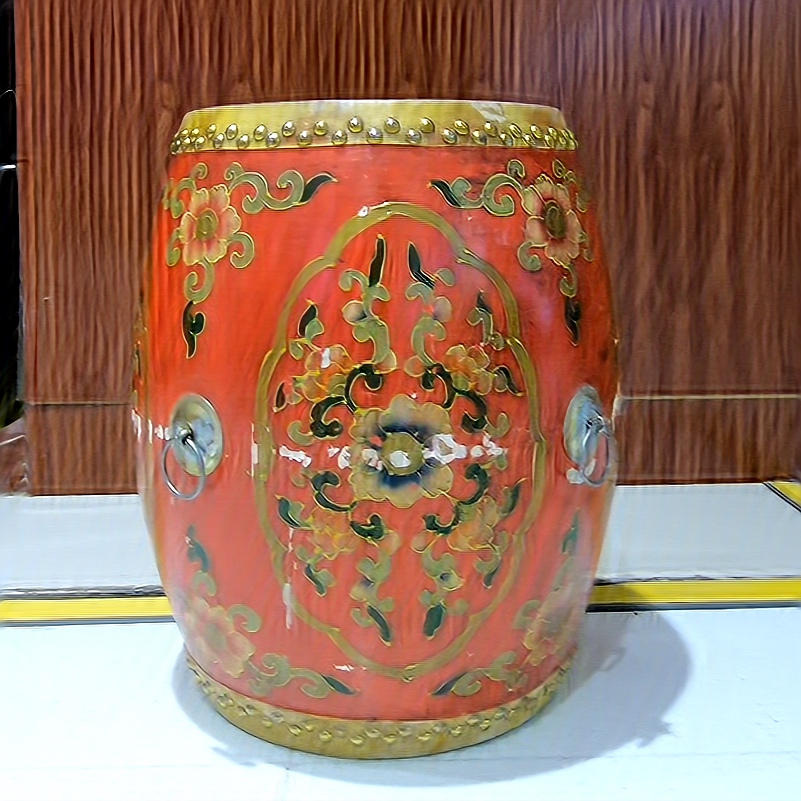}} &
   \multicolumn{3}{c}{\includegraphics[width=.16\linewidth]{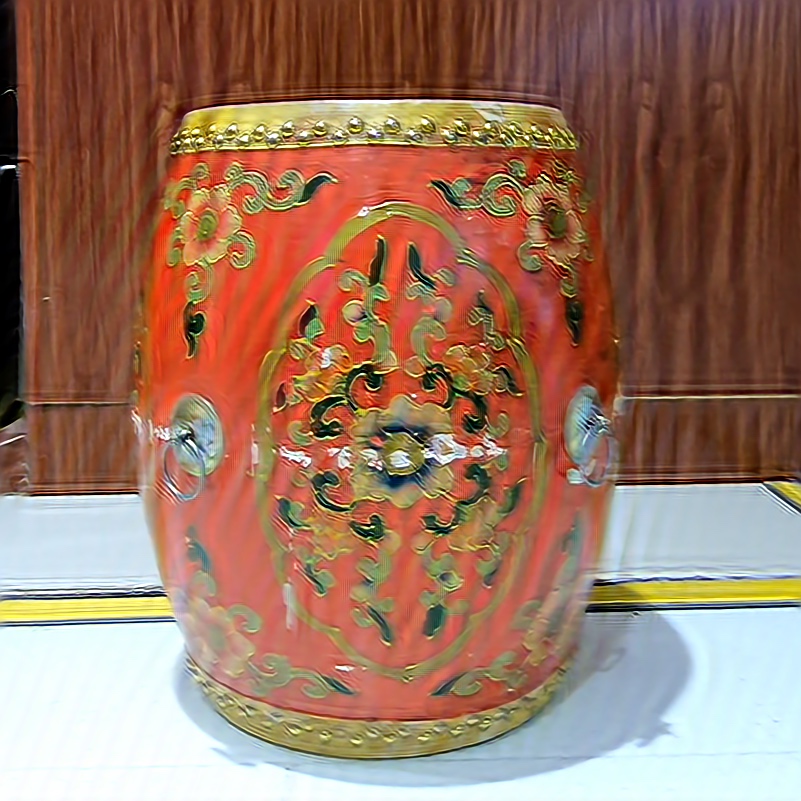}} &
   \multicolumn{3}{c}{\includegraphics[width=.16\linewidth]{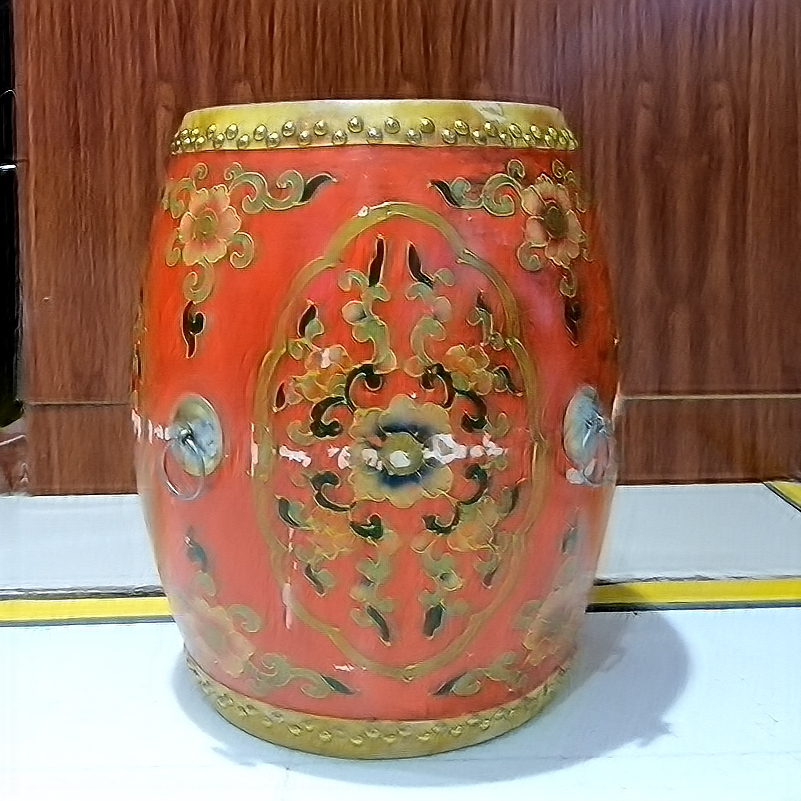}}\\

\multicolumn{3}{c}{Blurry images} &
\multicolumn{3}{c}{HL} &
\multicolumn{3}{c}{CSF} &
\multicolumn{3}{c}{IRCNN} &
\multicolumn{3}{c}{FDN} &
\multicolumn{3}{c}{CRCNet}
\end{tabular}
\caption{Visual comparison of deblurring results with state-of-art approaches on three real blurry images. From top to bottom are named as \texttt{lanttern}, \texttt{model} and \texttt{chair}.}
\label{fig_real}
\end{figure*}

\begin{table}[t]
\centering % used for centering table
\small
\begin{tabular}{c  c c  c} % centered columns (4 columns)
\hline %inserts double horizontal lines

Test image  & \texttt{lantern} & \texttt{model} & \texttt{chair}\\
\hline

Blurry      &3.3 &3.2 &3.5\\
HL         &8.6 &7.4 &8.3 \\
CSF        &8.0 &6.4 &7.7 \\
IRCNN      &8.3 &6.8 &5.1 \\
FDN        &8.4 &6.3 &6.3 \\
CRCNet     &\textbf{8.7} &\textbf{8.5} &\textbf{8.7}\\

\hline %inserts single line
\end{tabular}
\caption{Quantitative evaluations of different deblurring methods on real blurry images. Each number  notes the corresponding perceptual score proposed in~\cite{ma2017learning}.} % title of Table
\label{table:ssim_real} % is used to refer this table in the text

\end{table}

\begin{figure}[t]
\begin{center}
%\fbox{\rule{0pt}{2in} \rule{0.9\linewidth}{0pt}}
   \includegraphics[width=0.9\linewidth, trim={0pt 240pt 450pt 15pt}, clip]{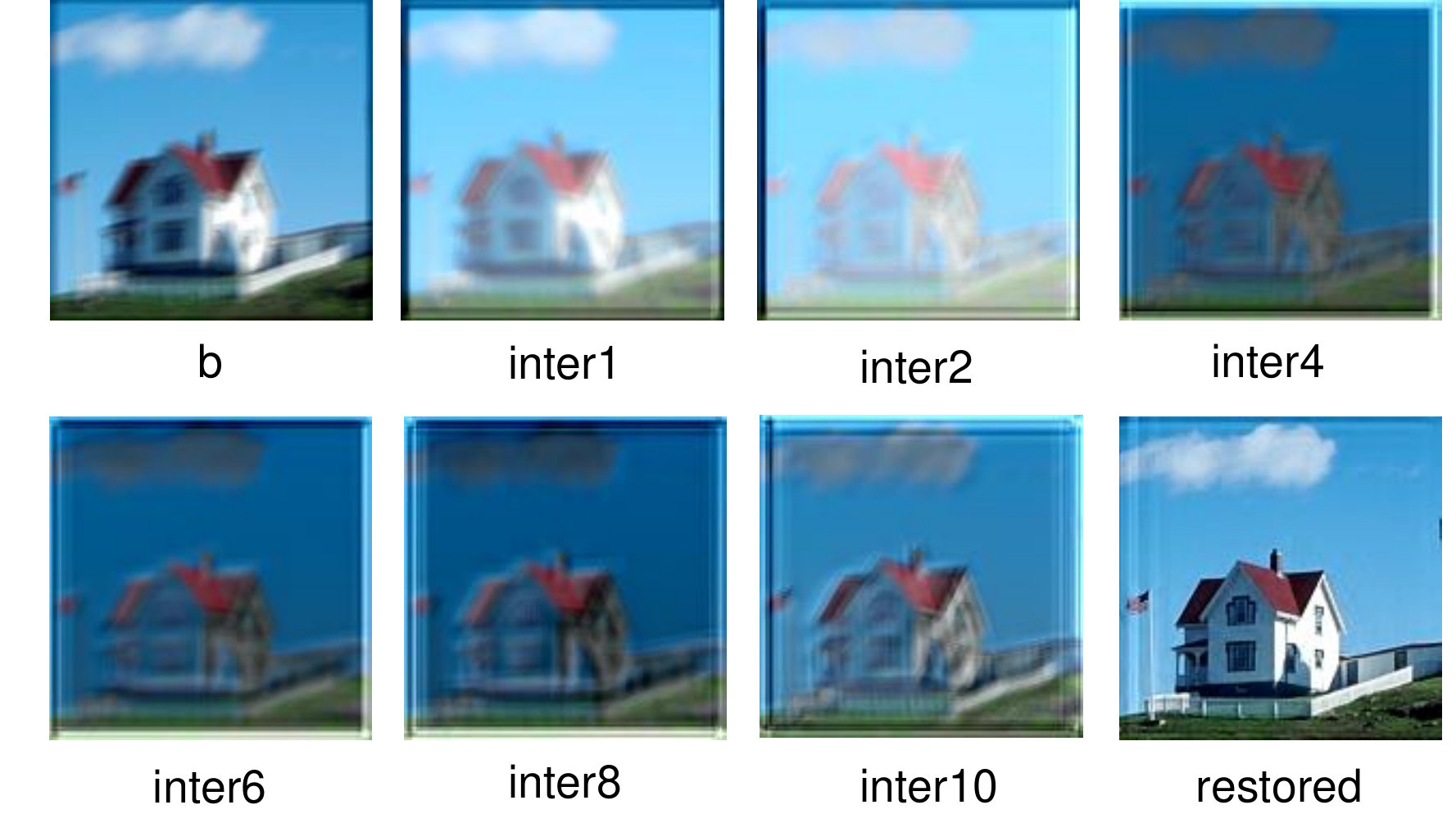}
\end{center}
   \caption{Intermediate samples of CRCNet. For visualization, deblurring is operated on the first channel of YCbCr and pixle values are normalized such that the maximum equals 1.}
\label{fig_inter}
\end{figure}

\subsection{Analysis to CRCNet}

A question beyond the superior performance is whether the effectiveness of CRCNet depends on our proposed concatenated residual (CR) architecture or just a trivial `universal approximator' relying on neural networks. To verify the contribution of CR structure, we give a discussion on relationship between IRD and CRCNet.

CRCNet plays a sequential nonlinear expansion of the iterative structure of IRD. 
Specifically, CRCNet realizes iteretive residues by several learned isolated \textit{conv/deconv} layers rather than thausands of iterations using fixed shared weights $k$ and $f_x$ in IRD (see Figure~\ref{fig:dp}). 
This iterative-to-sequential expansion enhances the flexibility and capacity of original method. In IRD algorithm, a large number of iterations are required for satisfactory deblurring quality; 
but in CRCNet, due to the powerful modeling capability of CNN, a very small number of layers can provide good restoration quality.

The iterative residual structure drives CRCNet to proccess images like IRD. To illustrate this point,
we visualized intermediate outputs $inter_i$.  
Figure~\ref{fig_inter} shows that deep outputs of CRCNet contain high-frequency oscilations along edges in the image. That fact  actually resembles IRD algorithm extracting high-freqency details after large amounts of iterations, as shown in Figure~\ref{fig2}.

We in this paper claim that deep CNN-based model for deconvolution should be equipped with specific architecture instead of plain CNN, and our proposed CRCNet is one of the potential effective architectures. 

% figure

% \paragraph{Convolutional Networks without CR Structure.}
% To show the effectiveness of CR structure, concatenation and residue components were knocked off the network described in Section 3, and compared with the complete CRCNet quantitatively.
% Table~\ref{table:ssim2} shows that both the network without concatenation and that without residue degenerate sharply. This fact shows the utility of CRCNet relies on our exquisitely designed structure.

% \begin{table}[h]
% \centering % used for centering table
% \begin{tabular}{c c c} % centered columns (4 columns)
% \hline %inserts double horizontal lines
% CRCNet & without concatenation & without residue\\  % inserts table
% %heading
% \hline
%  % inserts single horizontal line
%   % [1ex]

%    \textbf{0.8797} & 0.5985 &\\

% \hline %inserts single line
% \end{tabular}

% \caption{SSIM comparision on complete and partial CRCNets.} % title of Table
% \label{table:ssim2} % is used to refer this table in the text

% \end{table}

%As a conclusion, we expanded the traditional MMSE deconvolution into iterative residue series and designed a concatenated resudue network structure, which achieved high performance in non-blind image deblurring tasks.

\section{Conclusions}
In this paper, we proposed an effective deep architecture for deconvolution. 
By deriving the MMSE-based deconvolution solution, we first proposed an iterative residual deconvolution algorithm, which is simple yet effective. 
We further designed a concatenated residual convolutional network with the basic architecture of IRD algorithm. The restored results by CRCNet are more visually plausible compared with competing algorithms.
The success of CRCNet shows that deep CNN-based restoration architecture should borrow ideas from conventional methods.
In the future, we will develop more effective deep CNN-based restoration methods for other low level vision tasks.  

% \section*{Acknowledgments}
% This work is partially funded by NSFC (61472043) and NKRDPP (2017YFC1502505).
% Li Si-Yao extends special thanks to Wuyue Yu for offering images with good qualities during our experiments. We thank Wenyi Zeng and Ping Guo for constructive conversations.
% Q. Yin is the contact author.

%\appendix

%% The file named.bst is a bibliography style file for BibTeX 0.99c
\bibliographystyle{aaai}
\bibliography{crcdeconv2}
\end{document}